\documentclass[conference]{IEEEtran}
\IEEEoverridecommandlockouts
\usepackage{cite}
\usepackage{amsmath,amssymb,amsfonts}
\usepackage{graphicx}
\usepackage{textcomp}
\usepackage{xcolor}
\usepackage{booktabs}
\usepackage{makecell}
\usepackage{array}
\usepackage{cite}

\usepackage[margin=1in]{geometry}

\usepackage{amsmath}
\usepackage{amssymb}
\usepackage{amsfonts}
\usepackage{bm}
\usepackage{mathtools}

\usepackage{booktabs}
\usepackage{multirow}
\usepackage{array}

\usepackage{graphicx}
\usepackage{float}
\usepackage{xcolor}

\usepackage{algorithm}
\usepackage{algpseudocode}

\usepackage{hyperref}

\newcommand{\1}{\mathbf{1}}

\newcommand{\Bias}{\operatorname{Bias}}
\newcommand{\RMSE}{\operatorname{RMSE}}

\newcommand{\RCRG}{\operatorname{RCRG}}
\newcommand{\RBA}{\operatorname{RBA}}

\def\BibTeX{{\rm B\kern-.05em{\sc i\kern-.025em b}\kern-.08em
    T\kern-.1667em\lower.7ex\hbox{E}\kern-.125emX}}
\begin{document}

\title{Latent-Regime Bias Auditing \\ for Volatility Forecasting}

\author{
\IEEEauthorblockN{
Arthur Chagas,
Pedro Bento,
Yan Aquino,
Arthur Buzelin,\\
Wagner Meira Jr.,
Cristiano Arbex Valle
}
\IEEEauthorblockA{
\textit{Department of Computer Science}\\
\textit{Universidade Federal de Minas Gerais -- UFMG}\\
Belo Horizonte, Brazil\\
\{arthurchagas, pedro.bento, yanaquino, arthurbuzelin, meira, arbex\}@dcc.ufmg.br
}
}

\maketitle

\begin{abstract}
Volatility forecasts are commonly evaluated with aggregate accuracy metrics such as RMSE and MAE, but these metrics can hide conditional failures that matter for risk management. This paper proposes a model-agnostic audit framework for evaluating whether volatility forecasts remain reliable across latent market regimes. We learn time-series representations of market-state windows, cluster them into regimes using only training information, assign regimes out of sample, and compare aggregate forecast behavior with regime-conditional bias, tail-underprediction, and underprediction-sensitive economic losses. Applied to daily volatility forecasting across cryptocurrency and ETF assets, the audit shows that models with competitive aggregate accuracy can still exhibit substantial regime-specific bias and severe tail underprediction. The results suggest that volatility forecasting should be evaluated not only by average error, but also by where and how forecasts become unreliable. Our framework shifts forecast evaluation from asking which model is most accurate on average to identifying the market regimes in which apparently accurate forecasts fail conditionally. Reproducibility: \hyperlink{GitHub}{https://github.com/arthurchagas1/Latent-Regime-Bias-Auditing-for-Volatility-Forecasting}
\end{abstract}

\begin{IEEEkeywords}
volatility forecasting, latent market regimes, forecast reliability, regime-conditional bias, financial machine learning, time-series representation learning
\end{IEEEkeywords}

\section{Introduction}
\label{sec:introduction}

Volatility forecasts are most valuable precisely when market conditions become unstable. Risk limits, hedging decisions, volatility targeting, and portfolio allocation all depend on forecasts that remain reliable not only on average, but also in stressed market states. Yet volatility-forecasting models are still commonly evaluated through aggregate metrics such as RMSE and MAE. These metrics are necessary, but they compress away where errors occur. A forecast can look accurate overall while becoming systematically biased in the regimes where reliability matters most.

This paper argues that volatility forecasting should be evaluated as a conditional reliability problem. The relevant question is not only whether a model has low average error, but whether its errors remain stable across economically meaningful states\cite{christopher}. In risk-management settings, a model that underpredicts volatility during stress and compensates elsewhere is not merely noisy; it is conditionally unreliable\cite{elliott}. Standard aggregate evaluation can therefore obscure failures that are economically important.

We propose a model-agnostic audit framework for detecting such failures. The method learns time-series representations of market-state windows, reduces persistent asset identity through asset-normalization and adversarial representation learning, constructs within-asset latent regimes, and aligns these regimes by an economic stress score. The resulting regimes are not treated as universal market states or as forecasting targets. They are asset-relative audit partitions used to ask where an apparently accurate forecast becomes unreliable.

Empirically, we apply the framework to daily realized-volatility forecasts for cryptocurrency and ETF assets, auditing historical, HAR/IV, linear, tree-based, neural, and Transformer-style models. Results show that rankings differ across aggregate accuracy, conditional reliability, and underprediction-sensitive loss: strong average forecasters can still exhibit hidden regime bias, and more complex architectures do not necessarily improve tail reliability.

Our contributions are as follows:
\begin{enumerate}
    \item We frame volatility forecasting as an asset-relative, regime-conditional reliability auditing problem.

    \item We introduce model-agnostic diagnostics for hidden regime bias, tail underprediction, and underprediction-sensitive loss.

    \item We construct asset-relative regimes using normalized time-series embeddings, adversarial asset-invariance, within-asset clustering, and stress-score alignment.

    \item We audit a broad set of volatility forecasters, including HAR/IV, linear, tree-based, recurrent, convolutional, and Transformer-style models.

    \item We show that RMSE, conditional reliability, and economic-loss criteria can rank models differently.
\end{enumerate}

\begin{table*}[t]
\centering
\caption{Positioning relative to adjacent evaluation paradigms.}
\label{tab:positioning}
\scriptsize
\begin{tabular}{lccc}
\toprule
Approach & Groups/regimes & Used for prediction? & Main question \\
\midrule
Regime switching\cite{hamilton1989regime} & Model-implied & Yes & Which regime generated returns? \\
CPA tests\cite{giacomini2006tests} & Observed information sets & No & Does accuracy vary conditionally? \\
Subgroup auditing\cite{hebertjohnson2018multicalibration} & Predefined groups & No & Is performance fair/reliable by group? \\
\textbf{Ours} & \textbf{Learned time-series states} & \textbf{No} & \textbf{Where does forecast reliability fail?} \\
\bottomrule
\end{tabular}
\end{table*}

\section{Related Work}

Volatility forecasting has a long econometric tradition, including ARCH/GARCH models~\cite{engle1982arch,bollerslev1986garch}, realized-volatility models, and HAR-RV specifications~\cite{andersen2003modeling,andersen2007roughing,corsi2009har}. A related literature studies the predictive content of implied and option-implied volatility~\cite{christensen1998relation,busch2011volatility,patton2011volatility}. These approaches provide the main foundations for volatility prediction, but forecast evaluation is usually dominated by aggregate accuracy metrics. Our work differs by treating volatility forecasting as a conditional reliability problem: we ask whether models that look accurate on average remain reliable across latent market states.

Machine learning has expanded the volatility-forecasting toolkit through LSTM models~\cite{liu2019lstm}, intraday commonality features~\cite{zhang2024intradaycommonality}, GARCH-informed neural networks~\cite{xu2024garchinformed}, and language-model-based predictors such as Vola-BERT~\cite{nguyen2025volabert}. Recent reviews also document the broader use of AI for realized and implied volatility prediction~\cite{gunnarsson2024review}. Most of this work focuses on improving predictive accuracy or proposing new architectures. In contrast, we do not introduce another forecaster. We audit existing forecasters and test whether their errors become biased or tail-unreliable in specific market regimes.

Our regime construction is related to market-regime modeling and time-series representation learning. Classical work includes Markov-switching models~\cite{hamilton1989regime}, while recent representation methods include TS2Vec~\cite{yue2022ts2vec}, PatchTST~\cite{nie2023patchtst}, iTransformer~\cite{liu2024itransformer}, and DLinear-style baselines~\cite{zeng2023dlinear}. In finance, related studies examine recurring financial patterns~\cite{ibrain2024recurring} and contrastive asset embeddings~\cite{dolphin2024contrastiveasset}. These methods typically use regimes or representations to improve forecasting. Our use is different: regimes are not forecasting targets or model components, but asset-relative audit partitions used to locate conditional forecast failures.

This paper also connects to forecast comparison, conditional predictive ability, and reliability auditing. Standard forecast evaluation emphasizes global predictive accuracy~\cite{diebold1995comparing,white2000realitycheck,hansen2005spa}, while conditional predictive ability tests ask whether performance varies with observed information sets~\cite{giacomini2006tests}. In machine learning, calibration and subgroup reliability motivate evaluation beyond average performance~\cite{hebertjohnson2018multicalibration}. We bring this perspective to volatility forecasting by defining latent-regime diagnostics for worst-regime bias, \(\RBA_m^{\mathrm{stable}}\), \(\RCRG_m^{\mathrm{norm}}\), tail underprediction, and underprediction-sensitive economic loss.

Prior work improves average volatility forecasts using volatility features, machine learning, or regime structure. In contrast, we propose a model-agnostic audit that evaluates where accurate aggregate forecasts fail across regimes.

\begin{figure*}[t]
    \centering
    \includegraphics[width=\textwidth]{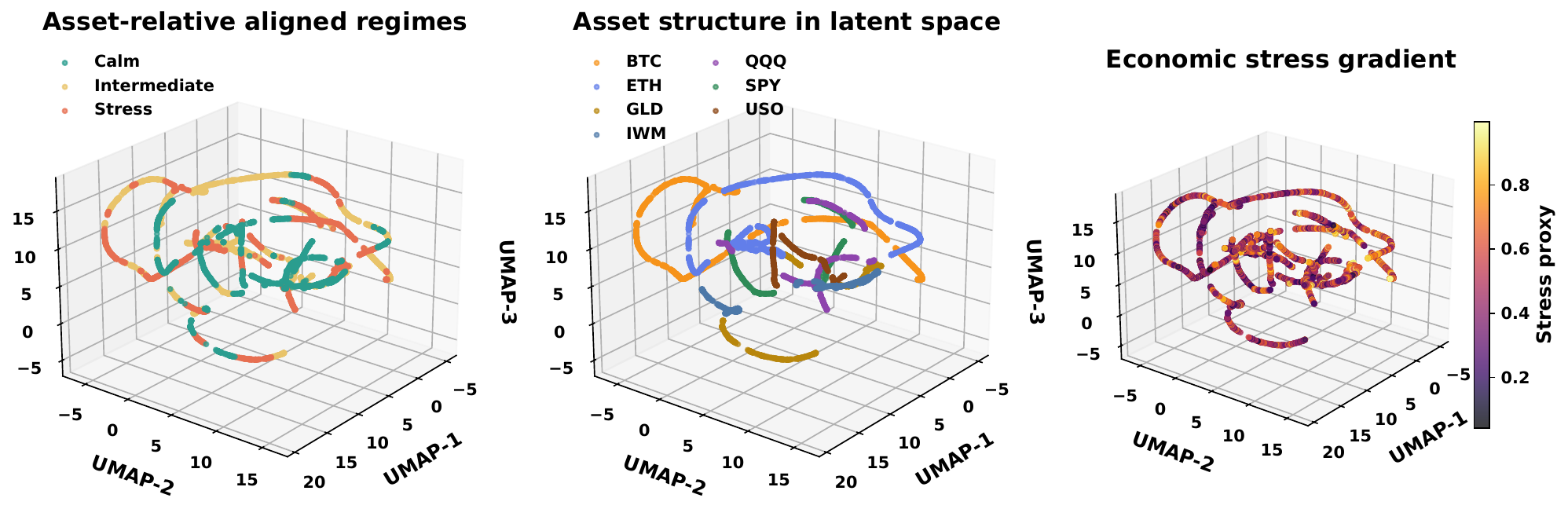}
    \caption{
    Asset-relative latent geometry after adversarial representation learning. The left panel shows the economically aligned asset-relative regimes, the middle panel colors the same latent space by asset, and the right panel shows the continuous stress score used to align within-asset regimes. The figure illustrates that the proposed representation no longer treats pooled clusters as universal market regimes; instead, it defines relative audit states that preserve market-state structure while explicitly exposing residual asset organization.
    }
    \label{fig:daily_latent_space}
\end{figure*}

\section{Methodology}
\label{sec:methodology}

We propose a model-agnostic audit that compares aggregate and regime-conditional volatility forecast reliability using asset-relative latent regimes.

\subsection{Information Set and Forecasting Target}
\label{subsec:information_target}

Let \(i\) index assets and \(t\) time. For each asset-time pair, we observe
\(
\bm{x}_{i,t}=f_{\mathrm{feat}}(\mathcal{D}_{i,t}),
\)
where \(\mathcal{D}_{i,t}\) contains only information available at time \(t\). Features include returns, range-based volatility, lagged realized volatility, implied volatility, volume, drawdown, and stress variables. For window length \(L\), the information set is
\(
\bm{X}_{i,t}^{(L)}
=
(\bm{x}_{i,t-L+1},\ldots,\bm{x}_{i,t}).
\)
All embeddings, regimes, and forecasts are constructed from training-fitted transformations and assigned out of sample. The target is future realized volatility over horizon \(h\):
\(
y_{i,t+h}
=
RV_{i,t}^{(h)}
=
\sqrt{
\sum_{j=1}^{h}
r_{i,t+j}^{2}
},
\)
where \(r_{i,t}\) is the log return.

\subsection{Asset-Normalized Representation Learning}
\label{subsec:asset_normalized_representation}

To reduce persistent asset-scale effects, each feature is normalized within asset using training-period robust statistics:
\[
\widetilde{x}_{i,t,j}
=
\frac{
x_{i,t,j}
-
\operatorname{median}_{s\in\mathcal{T}^{\mathrm{train}}_i}
x_{i,s,j}
}{
\operatorname{IQR}_{s\in\mathcal{T}^{\mathrm{train}}_i}
x_{i,s,j}
+
\epsilon
}.
\]
We then learn window embeddings with a TS2Vec-style contrastive encoder~\cite{yue2022ts2vec}:
\(
\bm{z}_{i,t}
=
\phi_{\theta}
\left(
\widetilde{\bm{X}}_{i,t}^{(L)}
\right)
\in\mathbb{R}^{d}.
\)
The encoder is trained to align temporally overlapping views and separate nonmatching windows. The embeddings are used only to define audit partitions, not as the main forecasting model.

To further reduce asset identity, we add an adversarial asset classifier with gradient reversal~\cite{ganin2016domain}. The classifier predicts the asset from \(\bm{z}_{i,t}\), while the encoder is trained to make this prediction difficult, preserving dynamic market-state information while discouraging asset-identifying structure.

\subsection{Asset-Relative Aligned Regimes}
\label{subsec:asset_relative_regimes}

Regimes are constructed within each asset. For asset \(i\), we fit \(K\)-means on training embeddings and assign held-out windows to the nearest training centroid:
\(
a_{i,t}\in\{1,\ldots,K\}.
\)
Because cluster labels are arbitrary across assets, we align them by a training-period stress score combining volatility, drawdown, absolute return, and stress-proxy information. Clusters are ordered by average stress score and labeled as
\(
c_{i,t}\in\{\text{calm},\text{intermediate},\text{stress}\}.
\)
Thus, regimes are asset-relative audit states: they need not be geometrically identical across assets, but they represent comparable within-asset stress levels. This avoids pooled clusters that mainly encode asset identity.

\subsection{Forecasting Models and Errors}
\label{subsec:forecasting_errors}

Let \(\mathcal{M}\) be the set of forecasting models. Each model \(m\in\mathcal{M}\) produces
\(
\widehat{y}_{m,i,t+h}
=
F_m(\bm{u}_{i,t}),
\)
where \(\bm{u}_{i,t}\) contains only information available at time \(t\), such as market features, realized-volatility lags, implied-volatility variables, embeddings, or asset indicators. The forecast error is
\(
e_{m,i,t+h}
=
\widehat{y}_{m,i,t+h}
-
y_{i,t+h}.
\)
Positive errors denote overprediction and negative errors denote underprediction. We report RMSE, MAE, and signed bias following standard forecast evaluation~\cite{diebold1995comparing,giacomini2006tests}.

\subsection{Regime-Conditional Reliability Audit}
\label{subsec:regime_reliability_audit}

For aligned regime \(k\), define
\(
\mathcal{I}_{k}
=
\{(i,t)\in\mathcal{I}: c_{i,t}=k\}.
\)
The regime-conditional bias is
\(
\widehat{\Bias}_{m,k}
=
\frac{1}{|\mathcal{I}_{k}|}
\sum_{(i,t)\in\mathcal{I}_{k}}
e_{m,i,t+h}.
\)
We also compute regime-specific RMSE, MAE, underprediction rates, and tail-underprediction rates, following conditional reliability evaluation~\cite{giacomini2006tests,hebertjohnson2018multicalibration}.

Let
\(
B_m^{\max}
=
\max_{k\in\mathcal{K}_{\mathrm{valid}}}
|\widehat{\Bias}_{m,k}|,
\)
where \(\mathcal{K}_{\mathrm{valid}}\) excludes regimes with insufficient test support. We define
\[
\RCRG_m^{\mathrm{norm}}
=
\frac{
B_m^{\max}
-
|\widehat{\Bias}_{m}|
}{
\RMSE_m+\epsilon
},
\]
and
\[
\RBA_m^{\mathrm{stable}}
=
\frac{
B_m^{\max}
}{
|\widehat{\Bias}_{m}|
+
\lambda_{\mathrm{stab}}\RMSE_m
+
\epsilon
}.
\]
These statistics quantify how much aggregate bias understates the largest asset-relative regime bias.

\subsection{Tail and Economic-Loss Evaluation}
\label{subsec:economic_loss}

Because volatility underprediction is costly for risk management, we use the asymmetric squared loss
\(
\ell_{\lambda}(e)
=
\lambda e^2 \1\{e<0\}
+
e^2\1\{e\ge 0\},
\space
\lambda>1.
\)
We also report a tail-weighted version for top-decile realized-volatility outcomes, plus mean shortfall and tail mean shortfall.

\subsection{Statistical Inference}
\label{subsec:statistical_inference}

We use block-bootstrap confidence intervals for regime-conditional bias, preserving local serial dependence in forecast errors. A regime is considered significantly biased when the interval for \(\widehat{\Bias}_{m,k}\) excludes zero. Bootstrap settings, seeds, and hyperparameters are provided in the repository.

\section{Experiments}
\label{sec:experiments}

This section describes the empirical design used to test whether volatility forecasting models remain reliable across asset-relative latent market states.

\subsection{Protocol}
\label{sec:protocol}

We evaluate whether volatility forecasts that are accurate on average remain reliable across asset-relative market states. The pipeline consists of five steps: constructing daily rolling windows from market, realized-volatility, implied-volatility, volume, drawdown, and stress-related features; normalizing features within each asset using training-period statistics; learning adversarial TS2Vec-style embeddings; obtaining within-asset latent regimes and aligning them by a training-period stress score; and evaluating held-out forecasts using aggregate accuracy, regime-conditional reliability, tail-underprediction, and underprediction-sensitive economic losses.

The analysis is conducted at daily frequency for horizons \(h\in\{1,5,10,21\}\), with \(h=5\) used as the main reporting horizon. We compare historical baselines, finance-native HAR/IV specifications, regularized linear models, tree-based and gradient-boosting models, recurrent and convolutional models, and Transformer-style baselines. The goal is not only to rank models by RMSE or MAE, but to test whether their errors remain stable across calm, intermediate, and stress regimes.

For each model, we report RMSE, MAE, signed bias, tail-underprediction, and economic-loss metrics, both in aggregate and within each aligned regime. Hidden conditional failures are summarized using \(\RCRG_m^{\mathrm{norm}}\) and \(\RBA_m^{\mathrm{stable}}\). Statistical uncertainty is assessed through block-bootstrap confidence intervals for regime-specific bias, and robustness checks vary the forecasting horizon and the regime-construction method.

\begin{figure*}[t]
\centering
\includegraphics[width=\textwidth]{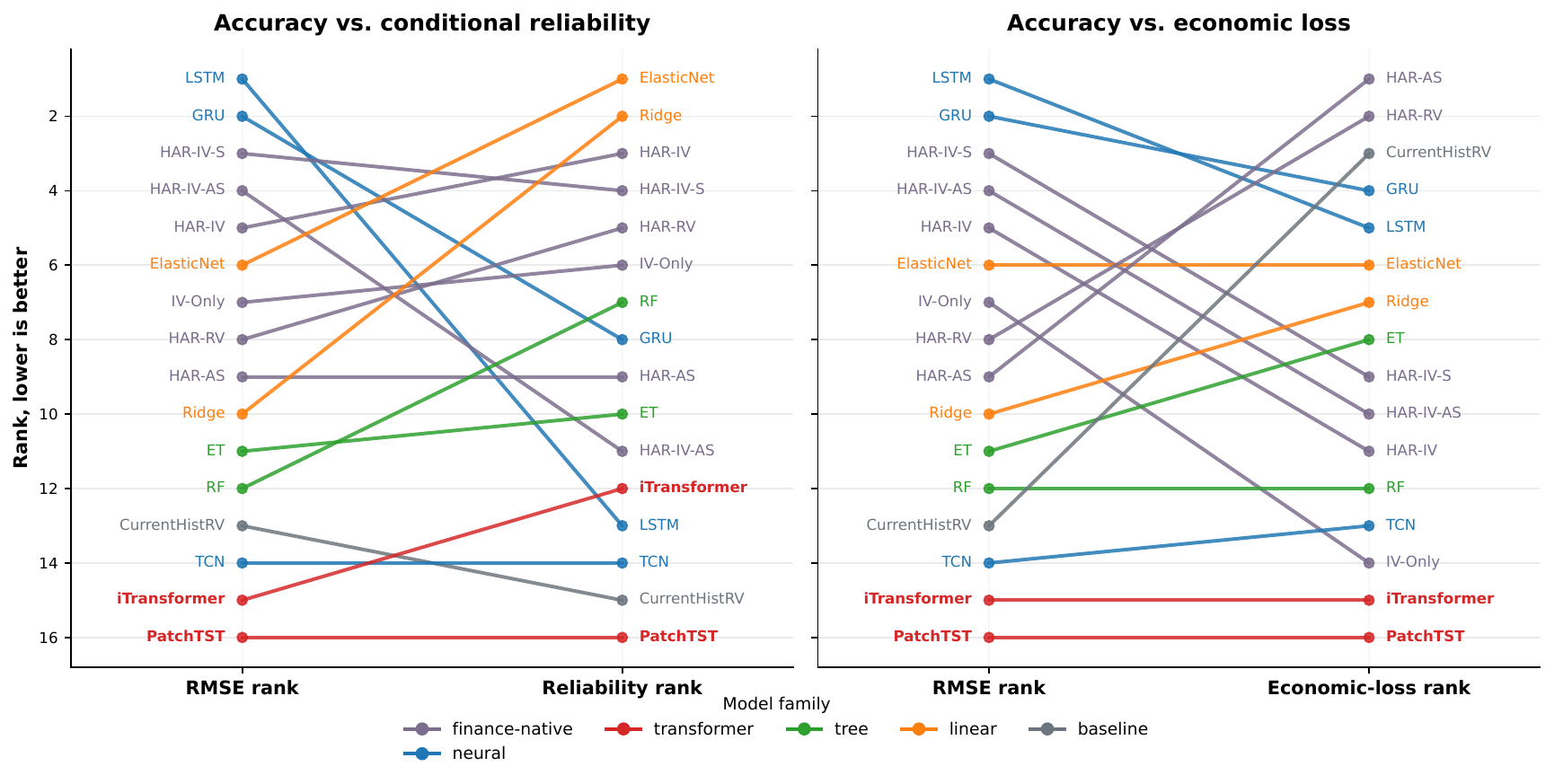}
\caption{
Metric-dependent model rankings. The left panel compares each model's aggregate RMSE rank with its conditional-reliability rank, computed from \(\RBA^{\mathrm{stable}}\). The right panel compares RMSE rank with the rank induced by tail-weighted asymmetric loss. Lower ranks are better. The rank shifts show that aggregate accuracy, regime-conditional reliability, and economic-risk criteria can lead to different assessments of the same volatility forecasting models. HAR-IV-S denotes HAR-RV with implied volatility and stress features; HAR-IV-AS denotes the asset-specific HAR-RV with implied volatility; ET denotes ExtraTrees; RF denotes RandomForest.
}
\label{fig:rank_shift}
\end{figure*}

\subsection{Data and Reproducibility Details}
\label{subsec:data_reproducibility}

We use daily data for BTC, ETH, SPY, QQQ, IWM, GLD, and USO from 2021-05-25 to 2026-03-07. OHLCV data come from \cite{binance_spot_klines,yfinance_docs}, and implied-volatility inputs from \cite{deribit_dvol_api,fred_vixcls}. Intraday records, when present, are aggregated to daily OHLCV using first open, maximum high, minimum low, last close, summed volume, and last available implied-volatility value. Each asset is split chronologically into train/validation/test periods using a \(70\%/15\%/15\%\) split.

The main specification uses \(L=64\)-day windows, embedding dimension \(d=64\), and \(K=3\) within-asset \(K\)-means regimes. Clusters are fit only on training embeddings, assigned out of sample by nearest centroid, and aligned as calm, intermediate, and stress using the average training percentile rank of realized volatility, absolute returns, high--low range, drawdown, and stress-proxy variables. Regime metrics are reported only when support is sufficient: \(n_{\min}=30\) and at least \(2\%\) of the evaluation subset. Inference uses \(B=500\) block-bootstrap replications with 10-day blocks and \(95\%\) confidence intervals.

\begin{table}[t]
\centering
\caption{Main experimental specification.}
\label{tab:experimental_spec}
\scriptsize
\setlength{\tabcolsep}{3.5pt}
\begin{tabular}{
@{}
>{\raggedright\arraybackslash}p{0.30\linewidth}
>{\raggedright\arraybackslash}p{0.62\linewidth}
@{}
}
\toprule
Item & Specification \\
\midrule
Assets & BTC, ETH, SPY, QQQ, IWM, GLD, USO \\
Period / frequency & 2021-05-25--2026-03-07, daily \\
Sources & OHLCV: \cite{binance_spot_klines,yfinance_docs}; IV: \cite{deribit_dvol_api,fred_vixcls} \\
Split & Chronological \(70\%/15\%/15\%\) per asset \\
Target / horizons & Future realized volatility, \(h\in\{1,5,10,21\}\); main \(h=5\) \\
Representation & \(L=64\), \(d=64\), TS2Vec-style encoder with asset adversary \\
Regimes & \(K=3\), within-asset \(K\)-means, stress-score alignment \\
Support / inference & \(n_{\min}=30\), \(2\%\) subset minimum; \(B=500\), 10-day block bootstrap \\
Seeds / tail set & 42, 43, 44; top decile of realized-volatility outcomes \\
\bottomrule
\end{tabular}
\end{table}

\subsection{Research Questions}
\label{sec:research_questions}

The experiments are organized around the following research questions.

\paragraph{RQ1: Do asset-relative regimes define economically usable audit states?}
We test whether the aligned regimes differ in interpretable market variables, including realized volatility, implied volatility, implied-to-realized spreads, drawdowns, volume, stress proxies, tail-event shares, and asset composition. This evaluates whether the regimes are useful audit partitions rather than only geometric clusters.

\paragraph{RQ2: Do aggregate metrics hide asset-relative regime bias?}
We compare aggregate signed bias with the largest aligned-regime bias using \(\RCRG_m^{\mathrm{norm}}\) and \(\RBA_m^{\mathrm{stable}}\). This tests whether models that appear reliable on average still exhibit systematic overprediction or underprediction in calm, intermediate, or stress states.

\paragraph{RQ3: Are reliability diagnostics distinct from aggregate accuracy?}
We examine whether regime-conditional reliability diagnostics provide information beyond RMSE and MAE. The goal is not to show that reliability is unrelated to accuracy, but to test whether models with similar aggregate accuracy can differ in conditional bias, tail-underprediction, and economic-loss exposure.

\paragraph{RQ4: Do model rankings change under economic-loss criteria?}
We evaluate whether the best models under aggregate RMSE remain preferable under underprediction-sensitive losses. This addresses whether volatility forecasts should be selected using a single average-error metric or evaluated jointly by accuracy, conditional reliability, and economic risk.

\paragraph{RQ5: Are the findings robust to horizon and representation choices?}
We test whether the main conclusions persist across forecasting horizons and alternative regime constructions, including plain asset-normalized embeddings, adversarial embeddings, and asset-residualized representations. This evaluates whether the observed conditional reliability gaps are stable features of the forecasting problem rather than artifacts of one specification.

\section{Results}
\label{sec:results}

This section presents the empirical evidence for the asset-relative latent-regime audit using daily volatility, adversarial TS2Vec embeddings, asset-relative aligned regimes, and out-of-sample forecasts from diverse forecasting models. Figure~\ref{fig:daily_latent_space} provides the latent-space overview, while the remaining tables and figures report the quantitative results.

\subsection{RQ1: Do Asset-Relative Regimes Define Economically Usable Audit States?}
\label{subsec:rq1_results}

Figure~\ref{fig:daily_latent_space} depicts the asset-relative latent audit space. The panels show the aligned calm, intermediate, and stress regimes, residual asset structure after adversarial learning, and the stress gradient used for alignment. The figure illustrates the construction: regimes are economically ordered, asset-relative audit states rather than pooled universal market regimes.

Table~\ref{tab:regime_economic_characterization} shows that the aligned regimes differ along interpretable financial dimensions. Calm periods have lower realized and implied volatility, smaller drawdowns, and no tail-event concentration. Intermediate and stress regimes exhibit higher volatility and larger tail-event shares, with the stress regime showing the largest drawdown. These patterns support the regimes as economically meaningful audit partitions.

\begin{table}[t]
\centering
\caption{
Economic characterization of the asset-relative aligned regimes in the held-out period. The regimes are ordered by the training-period stress score used for alignment. Tail Share denotes the share of top-decile five-day realized-volatility outcomes.
}
\label{tab:regime_economic_characterization}
\footnotesize
\setlength{\tabcolsep}{3.2pt}
\begin{tabular}{lcccccc}
\toprule
Regime
& \(n\)
& Crypto
& RV
& IV
& Drawdown
& Tail Share \\
\midrule
Calm
& 209
& 0.081
& 0.0255
& 25.25
& -0.016
& 0.000 \\
Intermediate
& 644
& 0.657
& 0.0492
& 44.73
& -0.055
& 0.121 \\
Stress
& 242
& 0.434
& 0.0473
& 44.09
& -0.114
& 0.116 \\
\bottomrule
\end{tabular}
\end{table}

Table~\ref{tab:asset_identity_suppression} shows that adversarial learning reduces linearly recoverable asset identity. Asset-prediction balanced accuracy falls from the plain embedding to the adversarial embedding and reaches chance level after residualization. This does not imply that asset information is fully removed, but it shows that the regime construction explicitly controls the asset-identity confound.

\begin{table}[t]
\centering
\caption{
Asset-identity predictability from alternative embeddings. Lower balanced accuracy indicates less linearly recoverable asset identity.
}
\label{tab:asset_identity_suppression}
\footnotesize
\setlength{\tabcolsep}{4pt}
\begin{tabular}{lccc}
\toprule
Embedding
& Accuracy
& Balanced Acc.
& Chance \\
\midrule
Plain
& 0.395
& 0.377
& 0.143 \\
Adversarial
& 0.317
& 0.291
& 0.143 \\
Adversarial + residualized
& 0.247
& 0.142
& 0.143 \\
\bottomrule
\end{tabular}
\end{table}

Overall, RQ1 is supported in the intended sense: the regimes are not universal market states, but economically ordered and statistically usable asset-relative audit partitions.

\subsection{RQ2: Do Aggregate Metrics Hide Asset-Relative Regime Bias?}
\label{subsec:rq2_results}

Table~\ref{tab:main_asset_relative_reliability} reports the main results at the five-day horizon. The best aggregate forecasters include recurrent models and finance-native HAR/IV benchmarks, confirming that the audit is evaluated against competitive baselines. However, low RMSE does not imply conditional reliability: several accurate models still exhibit regime-bias amplification and non-negligible underprediction-sensitive losses.

\begin{table}[t]
\centering
\caption{
Main asset-relative forecasting and reliability results at \(h=5\), ranked by aggregate RMSE. Tail Asym. is the tail-weighted asymmetric loss. Larger \(\RBA^{\mathrm{stable}}\) indicates stronger hidden conditional bias.
}
\label{tab:main_asset_relative_reliability}
\footnotesize
\setlength{\tabcolsep}{2.0pt}
\begin{tabular}{lccccc}
\toprule
Model
& RMSE
& MAE
& Bias
& \(\RBA^{\mathrm{stable}}\)
& Tail Asym. \\
\midrule
LSTM & 0.025 & 0.016 & -0.003 & 1.374 & 0.012 \\
GRU & 0.025 & 0.017 & -0.002 & 1.249 & 0.012 \\
HAR-IV-S & 0.025 & 0.016 & -0.003 & 1.131 & 0.013 \\
HAR-IV-AS & 0.025 & 0.016 & -0.001 & 1.351 & 0.014 \\
HAR-IV & 0.026 & 0.016 & -0.003 & 1.109 & 0.014 \\
ElasticNet & 0.026 & 0.017 & -0.002 & 0.734 & 0.013 \\
IV-Only & 0.026 & 0.016 & -0.003 & 1.199 & 0.014 \\
HAR-RV & 0.026 & 0.017 & 0.001 & 1.136 & 0.012 \\
HAR-AS & 0.026 & 0.018 & 0.001 & 1.292 & 0.012 \\
Ridge & 0.026 & 0.018 & -0.002 & 0.763 & 0.013 \\
ExtraTrees & 0.027 & 0.018 & 0.001 & 1.327 & 0.013 \\
RandomForest & 0.027 & 0.017 & -0.001 & 1.241 & 0.014 \\
CurrentHistRV & 0.032 & 0.022 & -0.001 & 2.420 & 0.012 \\
TCN & 0.033 & 0.024 & 0.001 & 1.968 & 0.014 \\
iTransformer & 0.034 & 0.024 & -0.005 & 1.362 & 0.020 \\
PatchTST & 0.036 & 0.026 & -0.000 & 3.490 & 0.021 \\
\bottomrule
\end{tabular}
\end{table}

The Transformer-style baselines are informative. In this daily setting, PatchTST and iTransformer do not dominate the strongest recurrent or finance-native models, and their larger tail-sensitive losses suggest weaker responsiveness to volatility extremes. This reinforces the main point: architectural complexity alone does not ensure conditional reliability.

\begin{figure}[t]
    \centering
    \includegraphics[width=\columnwidth]{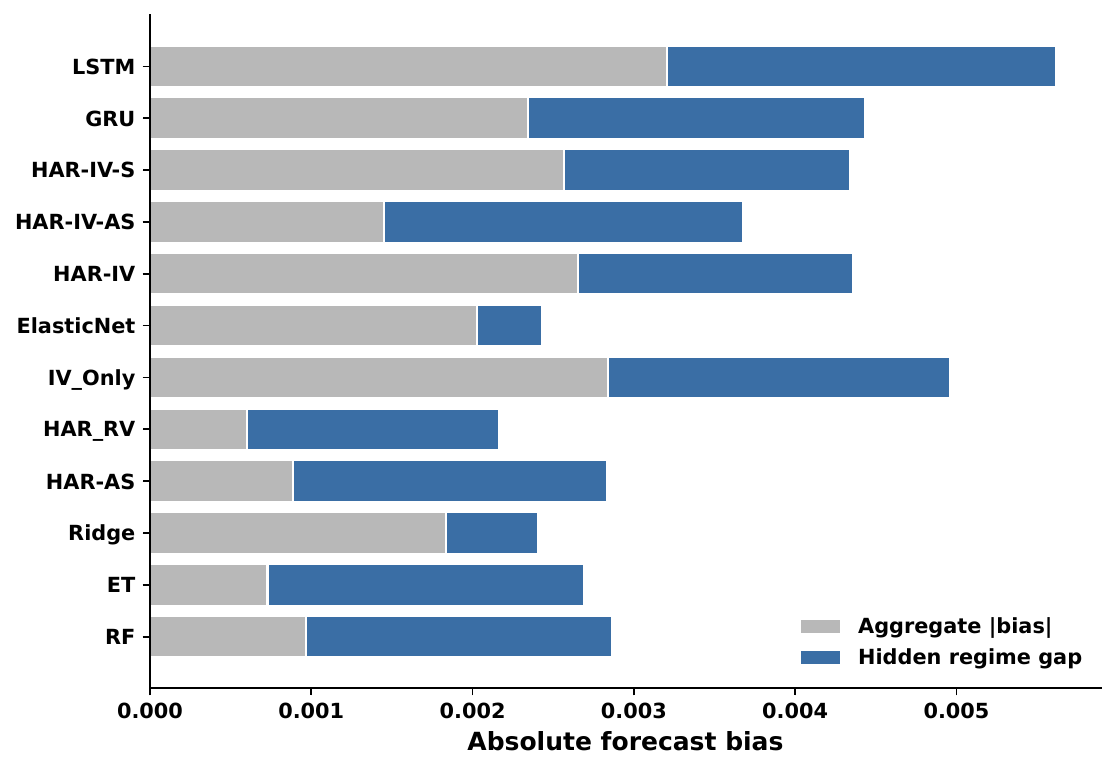}
    \caption{
    Hidden bias under asset-relative aligned regimes. Gray bars show absolute aggregate bias, while blue bars show the additional gap to the largest asset-relative regime bias. The full bar represents the worst conditional bias revealed by the audit.
    }
    \label{fig:hidden_bias_asset_relative}
\end{figure}

Figure~\ref{fig:hidden_bias_asset_relative} decomposes aggregate and regime-conditional bias. The gray segment shows absolute aggregate bias, while the blue segment shows the additional gap between aggregate bias and the largest asset-relative regime bias. The full bar therefore represents the worst conditional bias detected by the audit.

Table~\ref{tab:bootstrap_bias_summary} provides bootstrap evidence that these effects are not only point-estimate artifacts. Several models exhibit confidence intervals excluding zero in at least one aligned regime, indicating statistically supported conditional bias.

Overall, the results support RQ2: aggregate metrics can mask systematic forecast errors across calm, intermediate, and stress regimes. This is especially important for volatility forecasting, where underprediction in high-risk states is more costly than average error alone suggests.

\subsection{RQ3: Are Reliability Diagnostics Distinct from Aggregate Accuracy?}
\label{subsec:rq3_results}

The left panel of Figure~\ref{fig:rank_shift} compares each model's RMSE rank with its conditional-reliability rank. If aggregate accuracy and regime-conditional reliability were equivalent, the rank lines would remain nearly horizontal. Instead, several models shift substantially. Recurrent models rank best by RMSE but fall under the reliability criterion, while some linear and finance-native models become more favorable under conditional reliability.

This supports RQ3. The proposed diagnostics do not replace RMSE; they answer a different question. RMSE measures average error, whereas the asset-relative audit asks whether the error profile remains stable across economically aligned within-asset states. The results show that a model can be accurate on average while still displaying nontrivial conditional bias.

\begin{table}[t]
\centering
\caption{Block-bootstrap evidence for asset-relative conditional bias at $h=5$. For each model, the table reports the aligned regime with the largest absolute conditional bias, its bias, block-bootstrap confidence interval, exclusion of zero, and the number of bootstrap-supported nonzero-bias aligned regimes.}
\label{tab:bootstrap_bias_summary}
\scriptsize
\setlength{\tabcolsep}{2.3pt}
\begin{tabular}{lccccc}
\toprule
Model
& Worst Reg.
& Bias
& 95\% CI
& Worst Sig.
& Sig. Reg. \\
\midrule
LSTM & Intermediate & -0.006 & [-0.010, -0.001] & \textcolor{green!60!black}{Yes} & 1 \\
GRU & Intermediate & -0.004 & [-0.009, 0.000] & \textcolor{red!70!black}{No} & 0 \\
HAR-IV-S & Intermediate & -0.004 & [-0.009, 0.001] & \textcolor{red!70!black}{No} & 1 \\
HAR-IV-AS & Calm & 0.004 & [0.002, 0.005] & \textcolor{green!60!black}{Yes} & 1 \\
HAR-IV & Intermediate & -0.004 & [-0.009, 0.001] & \textcolor{red!70!black}{No} & 1 \\
ElasticNet & Intermediate & -0.002 & [-0.008, 0.003] & \textcolor{red!70!black}{No} & 0 \\
IV-Only & Intermediate & -0.005 & [-0.010, 0.000] & \textcolor{red!70!black}{No} & 1 \\
HAR-RV & Stress & 0.002 & [-0.004, 0.006] & \textcolor{red!70!black}{No} & 0 \\
HAR-AS & Stress & 0.003 & [-0.003, 0.007] & \textcolor{red!70!black}{No} & 0 \\
Ridge & Intermediate & -0.002 & [-0.008, 0.003] & \textcolor{red!70!black}{No} & 0 \\
ExtraTrees & Calm & 0.003 & [0.000, 0.005] & \textcolor{green!60!black}{Yes} & 1 \\
RandomForest & Calm & 0.003 & [0.000, 0.005] & \textcolor{green!60!black}{Yes} & 1 \\
TCN & Intermediate & 0.003 & [-0.004, 0.011] & \textcolor{red!70!black}{No} & 1 \\
iTransformer & Calm & 0.014 & [0.010, 0.018] & \textcolor{green!60!black}{Yes} & 2 \\
PatchTST & Calm & 0.014 & [0.006, 0.024] & \textcolor{green!60!black}{Yes} & 1 \\
\bottomrule
\end{tabular}
\end{table}

\subsection{RQ4: Do Model Rankings Change Under Economic-Loss Criteria?}
\label{subsec:rq4_results}

The right panel of Figure~\ref{fig:rank_shift} compares RMSE rank with the rank induced by tail-weighted asymmetric loss. The ranking changes again. Some models that are strong by aggregate RMSE become less attractive when the evaluation penalizes underprediction in high-volatility outcomes. Conversely, HAR-RV and asset-specific HAR-RV move upward under economic-loss ranking, indicating that simple finance-native benchmarks can remain competitive under risk-sensitive criteria.

Transformer-style baselines perform poorly under this criterion. This does not weaken the audit; rather, it illustrates why model selection should not be based on model class or architectural complexity alone. In this setting, the Transformer-style models are less responsive to volatility extremes and therefore less attractive under tail-sensitive loss.

These results support RQ4. Aggregate accuracy, conditional reliability, and economic risk do not collapse into a single ordering. Model selection for volatility forecasting should therefore consider not only RMSE, but also conditional bias and the economic cost of underprediction.

\subsection{RQ5: Are the Findings Robust to Representation Choices?}
\label{subsec:rq5_results}

Table~\ref{tab:variant_robustness} compares the main asset-relative adversarial construction with alternative regime constructions. The asset-relative adversarial variant reduces asset-regime dependence relative to the plain asset-normalized variant while preserving a clear reliability signal. The global residualized alternative also produces a reliability signal, but the main specification is preferable because it combines within-asset regime construction with explicit economic alignment.
\begin{table}[t]
\centering
\caption{
Robustness across regime-construction variants. The main specification is asset-relative adversarial. NMI and Cramer's \(V\) measure asset--regime dependence; median \(\RBA\) and median \(\RCRG\) summarize the reliability signal across models at the main horizon.
}
\label{tab:variant_robustness}
\scriptsize
\setlength{\tabcolsep}{2pt}
\begin{tabular}{lccccc}
\toprule
Variant
& NMI
& Cramer's \(V\)
& Med. \(\RBA\)
& Med. \(\RCRG\)
& Min. \(n\) \\
\midrule
Asset-relative adv.
& 0.305
& 0.644
& 2.123
& 0.075
& 206 \\
Asset-relative plain
& 0.348
& 0.695
& 2.095
& 0.090
& 307 \\
Global residual. adv.
& 0.270
& 0.629
& 2.063
& 0.102
& 226 \\
\bottomrule
\end{tabular}
\end{table}
The robustness evidence supports the interpretation that conditional reliability gaps are not artifacts of a single clustering choice. At the same time, the asset-dependence diagnostics show why the paper does not claim to discover universal market regimes. The more defensible conclusion is that asset-relative, economically aligned partitions provide useful audit states for detecting conditional forecast fragility.

\subsection{Summary}
Across the five research questions, the results support the central diagnostic claim of the paper: volatility forecasting models can appear competitive under aggregate metrics while remaining unreliable in specific asset-relative market states. The proposed audit complements standard forecast evaluation by identifying hidden conditional bias, tail underprediction, and economic-loss exposure that aggregate RMSE alone can obscure.

\section{Conclusion}
\label{sec:conclusion}

This paper introduces an asset-relative latent-regime audit for volatility forecasting, evaluating whether accurate average forecasts remain reliable across economically aligned within-asset regimes. The results show that aggregate accuracy can mask substantial regime-dependent failures.

The study has limitations. The regimes are sample-dependent audit partitions, not universal market states, and the empirical analysis is limited to daily realized-volatility targets in a modest cross-asset panel. Broader option data, longer samples, intraday targets, and alternative volatility measures may reveal further reliability failures. Future work can use these audits to design calibration and forecasting methods that directly reduce conditional bias in high-risk regimes.

\section*{Acknowledgements}

This work was partially funded by CNPq, Capes, Fapemig, IAIA-INCT on AI, INCT-TILDIAR, and the Brazilian Center on Algorithm Transparency and Reliable AI

{\footnotesize
\bibliography{bib}

@String{Computing = "Computing" }

@ArtifactSoftware{R,
    title = {R: A Language and Environment for Statistical Computing},
    author = {{R Core Team}},
    organization = {R Foundation for Statistical Computing},
    address = {Vienna, Austria},
    year = {2019},
    url = {https://www.R-project.org/},
}

@article{engle1982arch,
 ISSN = {00129682, 14680262},
 URL = {http://www.jstor.org/stable/1912773},
 author = {Robert F. Engle},
 journal = {Econometrica},
 number = {4},
 pages = {987--1007},
 publisher = {[Wiley, Econometric Society]},
 title = {Autoregressive Conditional Heteroscedasticity with Estimates of the Variance of United Kingdom Inflation},
 urldate = {2026-07-22},
 volume = {50},
 year = {1982}
}

@article{bollerslev1986garch,
title = {Generalized autoregressive conditional heteroskedasticity},
journal = {Journal of Econometrics},
volume = {31},
number = {3},
pages = {307-327},
year = {1986},
issn = {0304-4076},
doi = {https://doi.org/10.1016/0304-4076(86)90063-1},
url = {https://www.sciencedirect.com/science/article/pii/0304407686900631},
author = {Tim Bollerslev}
}

@article{andersen2003modeling,
 ISSN = {00129682, 14680262},
 URL = {http://www.jstor.org/stable/3082068},
 author = {Torben G. Andersen and Tim Bollerslev and Francis X. Diebold and Paul Labys},
 journal = {Econometrica},
 number = {2},
 pages = {579--625},
 publisher = {[Wiley, Econometric Society]},
 title = {Modeling and Forecasting Realized Volatility},
 urldate = {2026-05-08},
 volume = {71},
 year = {2003}
}

@article{andersen2007roughing,
 ISSN = {00346535, 15309142},
 URL = {http://www.jstor.org/stable/40043095},
 author = {Torben G. Andersen and Tim Bollerslev and Francis X. Diebold},
 journal = {The Review of Economics and Statistics},
 number = {4},
 pages = {701--720},
 publisher = {The MIT Press},
 title = {Roughing It up: Including Jump Components in the Measurement, Modeling, and Forecasting of Return Volatility},
 urldate = {2026-05-08},
 volume = {89},
 year = {2007}
}

@article{corsi2009har,
    author = {Corsi, Fulvio},
    title = {A Simple Approximate Long-Memory Model of Realized Volatility},
    journal = {Journal of Financial Econometrics},
    volume = {7},
    number = {2},
    pages = {174-196},
    year = {2009},
    month = {03},
    issn = {1479-8409},
    doi = {10.1093/jjfinec/nbp001},
    url = {https://doi.org/10.1093/jjfinec/nbp001},
    eprint = {https://academic.oup.com/jfec/article-pdf/7/2/174/2543795/nbp001.pdf},
}

@article{CHRISTENSEN1998relation,
title = {The relation between implied and realized volatility},
journal = {Journal of Financial Economics},
volume = {50},
number = {2},
pages = {125-150},
year = {1998},
issn = {0304-405X},
doi = {https://doi.org/10.1016/S0304-405X(98)00034-8},
url = {https://www.sciencedirect.com/science/article/pii/S0304405X98000348},
author = {B.J. Christensen and N.R. Prabhala}
}

@unknown{busch2011volatility,
author = {Busch, Thomas and Christensen, Bent and Nielsen, Morten},
year = {2007},
month = {01},
pages = {},
title = {The Role of Implied Volatility in Forecasting Future Realized Volatility and Jumps in Foreign Exchange, Stock, and Bond Markets},
journal = {Journal of Econometrics},
doi = {10.2139/ssrn.1148738}
}

@article{patton2011volatility,
title = {Volatility forecast comparison using imperfect volatility proxies},
journal = {Journal of Econometrics},
volume = {160},
number = {1},
pages = {246-256},
year = {2011},
note = {Realized Volatility},
issn = {0304-4076},
doi = {https://doi.org/10.1016/j.jeconom.2010.03.034},
url = {https://www.sciencedirect.com/science/article/pii/S030440761000076X},
author = {Andrew J. Patton}
}

@article{liu2019lstm,
author = {Liu, Yang},
title = {Novel volatility forecasting using deep learning–Long Short Term Memory Recurrent Neural Networks},
year = {2019},
issue_date = {Oct 2019},
publisher = {Pergamon Press, Inc.},
address = {USA},
volume = {132},
number = {C},
issn = {0957-4174},
url = {https://doi.org/10.1016/j.eswa.2019.04.038},
doi = {10.1016/j.eswa.2019.04.038},
journal = {Expert Syst. Appl.},
month = oct,
pages = {99–109},
numpages = {11}
}

@article{zhang2024intradaycommonality,
    author = {Zhang, Chao and Zhang, Yihuang and Cucuringu, Mihai and Qian, Zhongmin},
    title = {Volatility Forecasting with Machine Learning and Intraday Commonality*},
    journal = {Journal of Financial Econometrics},
    volume = {22},
    number = {2},
    pages = {492-530},
    year = {2024},
    month = {06},
    issn = {1479-8409},
    doi = {10.1093/jjfinec/nbad005},
    url = {https://doi.org/10.1093/jjfinec/nbad005},
    eprint = {https://academic.oup.com/jfec/article-pdf/22/2/492/57067139/nbad005.pdf},
}

@article{gunnarsson2024review,
title = {Prediction of realized volatility and implied volatility indices using AI and machine learning: A review},
journal = {International Review of Financial Analysis},
volume = {93},
pages = {103221},
year = {2024},
issn = {1057-5219},
doi = {https://doi.org/10.1016/j.irfa.2024.103221},
url = {https://www.sciencedirect.com/science/article/pii/S1057521924001534},
author = {Elias Søvik Gunnarsson and Håkon Ramon Isern and Aristidis Kaloudis and Morten Risstad and Benjamin Vigdel and Sjur Westgaard}
}

@inproceedings{xu2024garchinformed,
author = {Xu, Zeda and Liechty, John and Benthall, Sebastian and Skar-Gislinge, Nicholas and McComb, Christopher},
title = {GARCH-Informed Neural Networks for Volatility Prediction in Financial Markets},
year = {2024},
isbn = {9798400710810},
publisher = {Association for Computing Machinery},
address = {New York, NY, USA},
url = {https://doi.org/10.1145/3677052.3698600},
doi = {10.1145/3677052.3698600},
booktitle = {Proceedings of the 5th ACM International Conference on AI in Finance},
pages = {600–607},
numpages = {8},
location = {Brooklyn, NY, USA},
series = {ICAIF '24}
}

@inproceedings{nguyen2025volabert,
author = {Nguyen, Quoc Anh and Guo, Ce and Luk, Wayne},
title = {Repurposing Language Models for FX Volatility Forecasting: A Data-Efficient and Context-Aware Approach},
year = {2025},
isbn = {9798400722202},
publisher = {Association for Computing Machinery},
address = {New York, NY, USA},
url = {https://doi.org/10.1145/3768292.3770386},
doi = {10.1145/3768292.3770386},
booktitle = {Proceedings of the 6th ACM International Conference on AI in Finance},
pages = {465–473},
numpages = {9},
location = {
},
series = {ICAIF '25}
}

@inproceedings{yue2022ts2vec,
  title={TS2Vec: Towards Universal Representation of Time Series},
  author={Zhihan Yue and Yujing Wang and Juanyong Duan and Tianmeng Yang and Congrui Huang and Yu Tong and Bixiong Xu},
  booktitle={AAAI Conference on Artificial Intelligence},
  year={2021},
  url={https://api.semanticscholar.org/CorpusID:237497421}
}

@inproceedings{
nie2023patchtst,
title={A Time Series is Worth 64 Words:  Long-term Forecasting with Transformers},
author={Yuqi Nie and Nam H Nguyen and Phanwadee Sinthong and Jayant Kalagnanam},
booktitle={The Eleventh International Conference on Learning Representations },
year={2023},
url={https://openreview.net/forum?id=Jbdc0vTOcol}
}

@inproceedings{
liu2024itransformer,
title={iTransformer: Inverted Transformers Are Effective for Time Series Forecasting},
author={Yong Liu and Tengge Hu and Haoran Zhang and Haixu Wu and Shiyu Wang and Lintao Ma and Mingsheng Long},
booktitle={The Twelfth International Conference on Learning Representations},
year={2024},
url={https://openreview.net/forum?id=JePfAI8fah}
}

@inproceedings{zeng2023dlinear,
author = {Zeng, Ailing and Chen, Muxi and Zhang, Lei and Xu, Qiang},
title = {Are transformers effective for time series forecasting?},
year = {2023},
isbn = {978-1-57735-880-0},
publisher = {AAAI Press},
url = {https://doi.org/10.1609/aaai.v37i9.26317},
doi = {10.1609/aaai.v37i9.26317},
booktitle = {Proceedings of the Thirty-Seventh AAAI Conference on Artificial Intelligence and Thirty-Fifth Conference on Innovative Applications of Artificial Intelligence and Thirteenth Symposium on Educational Advances in Artificial Intelligence},
articleno = {1248},
numpages = {8},
series = {AAAI'23/IAAI'23/EAAI'23}
}

@article{hamilton1989regime,
 ISSN = {00129682, 14680262},
 URL = {http://www.jstor.org/stable/1912559},
 author = {James D. Hamilton},
 journal = {Econometrica},
 number = {2},
 pages = {357--384},
 publisher = {[Wiley, Econometric Society]},
 title = {A New Approach to the Economic Analysis of Nonstationary Time Series and the Business Cycle},
 urldate = {2026-07-22},
 volume = {57},
 year = {1989}
}

@inproceedings{ibrain2024recurring,
author = {Ibrain, \'{A}lvaro and Hern\'{a}ndez, Ver\'{o}nica and Peinado, Luis},
title = {Unveiling Recurring Financial Patterns: Novel unsupervised filtering algorithms for enhanced forecasting},
year = {2024},
isbn = {9798400710810},
publisher = {Association for Computing Machinery},
address = {New York, NY, USA},
url = {https://doi.org/10.1145/3677052.3698596},
doi = {10.1145/3677052.3698596},
booktitle = {Proceedings of the 5th ACM International Conference on AI in Finance},
pages = {406–418},
numpages = {13},
location = {Brooklyn, NY, USA},
series = {ICAIF '24}
}

@article{dolphin2024contrastiveasset,
  title={Stock embeddings: Learning distributed representations for financial assets},
  author={Dolphin, Rian and Smyth, Barry and Dong, Ruihai},
  journal={arXiv preprint arXiv:2202.08968},
  year={2022}
}

@article{diebold1995comparing,
 ISSN = {07350015},
 URL = {http://www.jstor.org/stable/1392185},
 author = {Francis X. Diebold and Roberto S. Mariano},
 journal = {Journal of Business \& Economic Statistics},
 number = {3},
 pages = {253--263},
 publisher = {[American Statistical Association, Taylor & Francis, Ltd.]},
 title = {Comparing Predictive Accuracy},
 urldate = {2026-05-08},
 volume = {13},
 year = {1995}
}

@article{white2000realitycheck,
 ISSN = {00129682, 14680262},
 URL = {http://www.jstor.org/stable/2999444},
 author = {Halbert White},
 journal = {Econometrica},
 number = {5},
 pages = {1097--1126},
 publisher = {[Wiley, Econometric Society]},
 title = {A Reality Check for Data Snooping},
 urldate = {2026-05-08},
 volume = {68},
 year = {2000}
}

@article{hansen2005spa,
 ISSN = {07350015},
 URL = {http://www.jstor.org/stable/27638834},
 author = {Peter Reinhard Hansen},
 journal = {Journal of Business \& Economic Statistics},
 number = {4},
 pages = {365--380},
 publisher = {[American Statistical Association, Taylor & Francis, Ltd.]},
 title = {A Test for Superior Predictive Ability},
 urldate = {2026-05-08},
 volume = {23},
 year = {2005}
}

@article{giacomini2006tests,
 ISSN = {00129682, 14680262},
 URL = {http://www.jstor.org/stable/4123083},
 author = {Raffaella Giacomini and Halbert White},
 journal = {Econometrica},
 number = {6},
 pages = {1545--1578},
 publisher = {[Wiley, Econometric Society]},
 title = {Tests of Conditional Predictive Ability},
 urldate = {2026-05-08},
 volume = {74},
 year = {2006}
}

@InProceedings{hebertjohnson2018multicalibration,
  title = 	 {Multicalibration: Calibration for the ({C}omputationally-Identifiable) Masses},
  author =       {Hebert-Johnson, Ursula and Kim, Michael and Reingold, Omer and Rothblum, Guy},
  booktitle = 	 {Proceedings of the 35th International Conference on Machine Learning},
  pages = 	 {1939--1948},
  year = 	 {2018},
  editor = 	 {Dy, Jennifer and Krause, Andreas},
  volume = 	 {80},
  series = 	 {Proceedings of Machine Learning Research},
  month = 	 {10--15 Jul},
  publisher =    {PMLR},
  url = 	 {https://proceedings.mlr.press/v80/hebert-johnson18a.html}
}

@article{ganin2016domain,
author = {Gani, Yaroslav and Ustinova, Evgeniya and Ajakan, Hana and Germain, Pascal and Larochelle, Hugo and Laviolette, Francois and Marchand, Mario and Lempitsky, Victor},
year = {2015},
month = {05},
pages = {},
title = {Domain-Adversarial Training of Neural Networks},
journal = {Journal of Machine Learning Research}
}

@misc{binance_spot_klines,
  author       = {{Binance}},
  title        = {{Spot API Documentation: Kline/Candlestick Data}},
  year         = {2026},
  howpublished = {\url{https://developers.binance.com/docs/binance-spot-api-docs/rest-api/market-data-endpoints}},
  note         = {Accessed: 2026-05-10}
}

@misc{deribit_dvol_api,
  author       = {{Deribit}},
  title        = {{Deribit API Documentation: public/get\_volatility\_index\_data}},
  year         = {2026},
  howpublished = {\url{https://docs.deribit.com/api-reference/market-data/public-get_volatility_index_data}},
  note         = {Accessed: 2026-05-10}
}

@misc{yfinance_docs,
  author       = {Aroussi, Ran},
  title        = {{yfinance: Download Market Data from Yahoo! Finance's API}},
  year         = {2026},
  howpublished = {\url{https://ranaroussi.github.io/yfinance/}},
  note         = {Accessed: 2026-05-10}
}

@misc{fred_vixcls,
  author       = {{Chicago Board Options Exchange}},
  title        = {{CBOE Volatility Index}},
  year         = {2026},
  howpublished = {Retrieved from FRED, Federal Reserve Bank of St. Louis},
  url          = {https://fred.stlouisfed.org/},
  note         = {Accessed: 2026-05-10}
}

@article{christopher,
 ISSN = {00206598, 14682354},
 URL = {http://www.jstor.org/stable/2527341},
 author = {Peter F. Christoffersen},
 journal = {International Economic Review},
 number = {4},
 pages = {841--862},
 publisher = {[Economics Department of the University of Pennsylvania, Wiley, Institute of Social and Economic Research, Osaka University]},
 title = {Evaluating Interval Forecasts},
 urldate = {2026-05-15},
 volume = {39},
 year = {1998}
}

@article{elliott,
 ISSN = {00346527, 1467937X},
 URL = {http://www.jstor.org/stable/3700702},
 author = {Graham Elliott and Ivana Komunjer and Allan Timmermann},
 journal = {The Review of Economic Studies},
 number = {4},
 pages = {1107--1125},
 publisher = {[Oxford University Press, Review of Economic Studies, Ltd.]},
 title = {Estimation and Testing of Forecast Rationality under Flexible Loss},
 urldate = {2026-05-15},
 volume = {72},
 year = {2005}
}
\bibliographystyle{IEEEtran}
}

\end{document}